\pdfoutput=1
\documentclass[10pt,twocolumn,letterpaper]{article}

\usepackage[margin=0.75in,columnsep=0.25in]{geometry}
\usepackage[T1]{fontenc}
\usepackage[utf8]{inputenc}
\usepackage{times}
\usepackage{amsmath,amssymb}
\usepackage{microtype}
\usepackage{graphicx}
\usepackage{booktabs}
\usepackage{adjustbox}
\usepackage{caption}
\usepackage{algorithm}
\usepackage{algpseudocode}
\usepackage[numbers,sort&compress]{natbib}
\usepackage{url}
\usepackage{hyperref}

\title{Boosting Transferable Adversarial Attacks against Deep Reinforcement Learning}

\author{%
  Zexin Li\textsuperscript{1}\quad
  Ruili Yao\textsuperscript{2}\quad
  Yiming Zeng\textsuperscript{3}\quad
  Xiaoxue Gao\textsuperscript{4}\thanks{Corresponding author. Assistant Professor, School of Artificial Intelligence, The Chinese University of Hong Kong, Shenzhen.}\\[0.6em]
  {\small \textsuperscript{1}Nanyang Technological University}\\
  {\small \textsuperscript{2}University of California, Riverside}\quad
  {\small \textsuperscript{3}University of Connecticut}\\
  {\small \textsuperscript{4}School of Artificial Intelligence,}\\
  {\small The Chinese University of Hong Kong, Shenzhen}\\[0.3em]
  {\small \texttt{zexin.li@ntu.edu.sg}}
}
\date{}
\hypersetup{
  hidelinks,
  pdftitle={Boosting Transferable Adversarial Attacks against Deep Reinforcement Learning},
  pdfauthor={Zexin Li, Ruili Yao, Yiming Zeng, Xiaoxue Gao}
}

\begin{document}

\maketitle

\begin{abstract}
Most adversarial attacks on deep reinforcement learning (DRL) assume white-box access to the victim policy, which rarely holds in practice. This paper studies \emph{transfer-based black-box} attacks on DRL: the attacker crafts observation perturbations on a white-box surrogate agent and feeds them to an unknown victim. We formulate the attack as return minimization under a per-step perturbation budget. We first show that transplanting transferable image-classification attacks (FGSM, MI-FGSM, and NI-FGSM) with a per-step objective yields perturbations that transfer but are no stronger than random noise of the same budget. We then propose a \emph{trajectory-level} attack that optimizes a sequence of perturbations over a receding horizon through a differentiable model of the environment and a temperature-smoothed surrogate policy, with the same optimizers. On CartPole-v1 with ten DQN and DDQN agents and 100 surrogate--victim pairs, the trajectory-level attack outperforms per-step attacks and random noise in the white-box, cross-model, and cross-algorithm settings.
\end{abstract}

\section{Introduction}
Deep reinforcement learning (DRL) has achieved remarkable success in learning complex control tasks and adapting to dynamic environments~\cite{mnih2015human,peng2018deepmimic,kahn2018self}. However, DRL policies inherit the adversarial fragility of deep neural networks: small, carefully crafted perturbations of the observations can steer an agent into catastrophic actions~\cite{huang2017adversarial,lin2017tactics,zhang2020robust}. Such vulnerabilities have been demonstrated across modalities and tasks, including face recognition~\cite{li2023siblingattack}, speech recognition~\cite{gao2024transferable}, dialogue generation~\cite{li2023whitebox}, the efficiency of dynamic neural networks and large language models~\cite{chen2023dynamicacl,chen2026naturalsloth}, backdoors in vision-language models~\cite{cheng2026vlmsystrojan}, and policy manipulation in multi-robot reinforcement learning~\cite{nikkhoo2023pimbotiros,li2026pimbot}.

Most attacks on DRL assume white-box access to the victim's parameters~\cite{huang2017adversarial,lin2017tactics,zhang2020robust,sun2022strongest}. In practice, a deployed policy is usually inaccessible, so the attacker has to operate in a \emph{black-box} setting~\cite{pantransferable,huang2017adversarial}. A practical strategy is to craft \emph{transferable} adversarial examples on a white-box surrogate model and apply them to the victim. This strategy has proven highly effective for image classification~\cite{dong2018boosting,lin2020nesterov,li2023siblingattack} and speech recognition~\cite{gao2024transferable}, and has also been demonstrated in DRL across policies and algorithms~\cite{huang2017adversarial,pantransferable}. In particular, prior work identifies shared adversarial features across MDPs~\cite{korkmaz2022shared} and shows that adversarial training can leave policies vulnerable to natural, black-box perturbations~\cite{korkmaz2023redefining}. Building on this evidence, we study whether explicitly optimizing future state trajectories improves transfer over per-step attack objectives.

A natural first step is to apply strong transferable attacks from image classification independently at every time step. We find that, on CartPole, such perturbations transfer across seeds and algorithms but are no more damaging than random noise at the corners of the same budget. The per-step objective is myopic: it only targets the current action, whereas a well-trained policy quickly corrects a single wrong action.

We therefore lift the attack objective from single actions to trajectories. Our attack optimizes a sequence of perturbations over a receding horizon so that the surrogate agent, rolled out through a differentiable environment model, is driven toward failure. We solve this problem with the momentum and Nesterov optimizers of MI-FGSM and NI-FGSM, and we smooth the surrogate policy with a temperature to avoid overfitting to its exact decision boundary.

Our contributions are: (1)~a return-minimization formulation of transfer attacks on DRL that explains why per-step relaxations fall short (\S\ref{sec:formulation}); (2)~a trajectory-level attack that unifies FGSM, MI-FGSM, and NI-FGSM under a receding-horizon objective (\S\ref{sec:traj}); and (3)~an evaluation of white-box, cross-model, and cross-algorithm transfer over ten DQN/DDQN agents and 100 surrogate--victim pairs (\S\ref{sec:exp}).

\section{Related Work}
\textbf{Transferable adversarial attacks.}
Transferability can be improved with iterative attacks~\cite{kurakin2016adversarial,madry2017towards}, momentum~\cite{dong2018boosting}, Nesterov look-ahead~\cite{lin2020nesterov}, variance tuning~\cite{wang2021enhancing}, and input diversity~\cite{xie2019improving}. They have also been tailored to face recognition~\cite{li2023siblingattack} and speech recognition~\cite{gao2024transferable}. Beyond accuracy, adversarial inputs can degrade model efficiency~\cite{chen2023dynamicacl,chen2026naturalsloth}, manipulate text generation~\cite{li2023whitebox}, or implant stealthy backdoors~\cite{cheng2026vlmsystrojan,guo2022backdoor}.

\textbf{Adversarial attacks on DRL.}
\citet{huang2017adversarial} first showed that FGSM~\cite{goodfellow2014explaining} perturbations degrade DRL policies in Atari games~\cite{bellemare2013arcade} and briefly examined transfer across policies and algorithms. \citet{lin2017tactics} proposed the strategically timed attack and the enchanting attack, where the latter uses a learned dynamics model to plan a sequence of target actions. Optimal white-box adversaries can be learned with reinforcement learning~\cite{zhang2020robust,sun2022strongest}, and domain randomization improves transfer~\cite{pantransferable}. In multi-agent settings, PIMbot manipulates the policies and incentives of robot teams~\cite{nikkhoo2023pimbotiros,li2026pimbot}. \citet{korkmaz2022shared} shows that sensitive directions can transfer across states and MDPs, including to adversarially trained policies. \citet{korkmaz2023redefining} further demonstrates that natural observation transformations can undermine nominally robust policies in a black-box setting. Our focus is on optimizing temporally coordinated perturbations through a differentiable dynamics model and evaluating their transfer across seeds and algorithms.

\textbf{Diagnosing and evaluating robustness.}
\citet{korkmaz2023detecting} detect adversarial directions using local quadratic approximations of the policy loss. \citet{korkmaz2024diagnosing} analyzes how unstable directions evolve across states and time, revealing limitations of robust training. \citet{korkmaz2026counterfactuality} shows that enforcing local invariance can impair counterfactual value learning. These findings motivate evaluating defenses beyond the particular perturbations used during training; our experiments examine attack effectiveness rather than establishing a robustness guarantee.

\textbf{Sequence-level objectives.}
In reinforcement learning for language-model reasoning, SPEAR~\cite{li2026spear} uses order-aware alignment of symbolic reasoning trajectories to provide process rewards for distillation. The connection to our work is the use of sequence structure in the objective; SPEAR studies reasoning distillation, whereas we optimize observation perturbations for adversarial control.

\section{Method}
\label{sec:method}

\begin{figure*}[t]
    \centering
    \includegraphics[width=\textwidth]{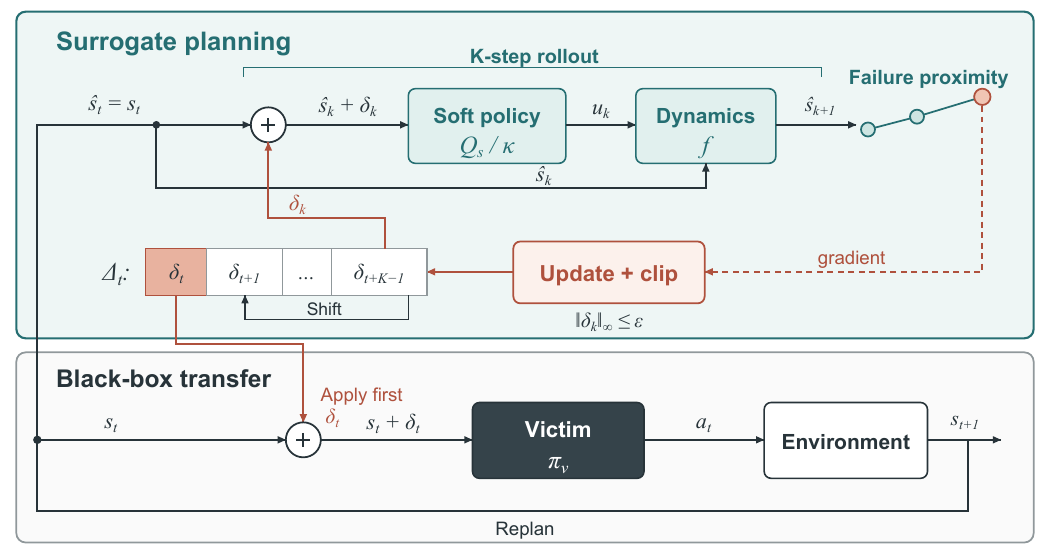}
    \caption{Trajectory-level transfer attack. The attacker optimizes a bounded perturbation sequence through a temperature-smoothed surrogate and differentiable dynamics to maximize predicted failure proximity. Only the first perturbation is applied to the victim's observation; the remainder warm-starts the next plan. The environment evolves from the true state. Dashed arrows denote gradients; no victim parameters or queries are used during crafting.}
    \label{fig:threat}
\end{figure*}

\subsection{Problem Formulation}
\label{sec:formulation}
We model the task as a Markov decision process $\mathcal{M}=(\mathcal{S},\mathcal{A},P,r,\gamma)$. The victim is a value-based agent with Q-network $Q_v(\cdot,\cdot;\theta_v)$ and greedy policy $\pi_v(s)=\arg\max_{a}Q_v(s,a;\theta_v)$. At each step $t$, the attacker observes the true state $s_t$ and injects a perturbation $\delta_t$ into the victim's observation (Figure~\ref{fig:threat}), so that the victim acts on $\tilde{s}_t=s_t+\delta_t$ while the environment evolves according to the true state. The attacker aims to minimize the victim's expected return under a per-step $\ell_\infty$ budget $\epsilon$:
\begin{equation}
\label{eq:goal}
\begin{aligned}
\min_{\{\delta_t\}_{t=0}^{H}} \;& J(\delta)=\mathbb{E}\Big[\textstyle\sum_{t=0}^{H} \gamma^{t} r(s_t,a_t)\Big] \\
\text{s.t.}\;\; & a_t=\pi_v(s_t+\delta_t),\;\; \|\delta_t\|_\infty\le\epsilon, \\
& s_{t+1}\sim P(\cdot\mid s_t,a_t),
\end{aligned}
\end{equation}
where $H$ is the episode horizon. In the black-box setting, $\theta_v$ is unknown and the attacker cannot query $\pi_v$ while crafting perturbations. It trains a surrogate Q-network $Q_s(\cdot,\cdot;\theta_s)$ in the same environment, possibly with a different seed or algorithm, and uses it as a stand-in for $Q_v$. Even with a surrogate, Eq.~\eqref{eq:goal} is hard to optimize: the return is a non-differentiable function of a long sequence of discrete actions and stochastic transitions.

\textbf{Per-step relaxation.} The standard relaxation, which we use as a baseline, optimizes each perturbation independently. We choose a lure action $a^\dagger$ whose persistent execution drives the system toward failure (for CartPole, always pushing the cart in the same direction), and at each step we solve
\begin{equation}
\label{eq:surrogate}
\delta_t^{*}=\arg\max_{\|\delta\|_\infty\le\epsilon}\; \log\operatorname{softmax}\big(Q_s(s_t+\delta,\cdot)\big)_{a^\dagger}
\end{equation}
with FGSM~\cite{goodfellow2014explaining}, MI-FGSM~\cite{dong2018boosting}, or NI-FGSM~\cite{lin2020nesterov}. This objective is myopic: one wrong action changes the state only slightly and the victim's next decision can undo it, so a per-step attack must succeed at almost every step, which is hard under a small budget and harder after transfer.

\subsection{Trajectory-Level Transfer Attack}
\label{sec:traj}
We instead optimize perturbations with respect to their effect on the future trajectory. We assume the attacker has a differentiable model $f$ of the environment dynamics, $\hat{s}_{k+1}=f(\hat{s}_k,u_k)$, where $u_k$ is a continuous relaxation of the action. For CartPole, $f$ is the known physical model; in general, it can be learned from interaction data, as in the enchanting attack~\cite{lin2017tactics}. To make the rollout differentiable, we replace the surrogate's greedy action with a temperature-smoothed policy
\begin{equation}
\label{eq:relax}
\tilde{\pi}_s(a\mid x)=\operatorname{softmax}\big(Q_s(x,\cdot)/\kappa\big)_a,
\end{equation}
and apply the expected action $u_k=\mathbb{E}_{a\sim\tilde{\pi}_s(\cdot\mid\hat{s}_k+\delta_k)}[a]$ (for CartPole, the expected force). At step $t$, the attacker optimizes a sequence of $K$ perturbations $\Delta_t=(\delta_{t},\dots,\delta_{t+K-1})$ to maximize the proximity of the predicted states to the failure region:
\begin{equation}
\label{eq:traj}
\max_{\|\delta_{k}\|_\infty\le\epsilon}\;\mathcal{L}_{\mathrm{traj}}(\Delta_t)=\sum_{k=1}^{K}\phi(\hat{s}_{t+k}),\quad \hat{s}_t=s_t,
\end{equation}
where $\phi(s)=\max\{(\theta/\theta_{\max})^2,(x/x_{\max})^2\}$ measures how close the pole angle $\theta$ and the cart position $x$ are to their termination thresholds. Following model-predictive control, only $\delta_t$ is applied, and the remaining perturbations warm-start the optimization at the next step.

We solve Eq.~\eqref{eq:traj} with the same operators as in the per-step case, applied to the whole sequence $\Delta_t$. With step size $\alpha=\epsilon/T$ and momentum decay $\mu$, MI-FGSM accumulates an $\ell_1$-normalized gradient momentum,
\begin{equation}
\label{eq:mi}
g_{j+1}=\mu\, g_j+\frac{\nabla_\Delta\mathcal{L}_{\mathrm{traj}}(\hat{\Delta}_j)}{\|\nabla_\Delta\mathcal{L}_{\mathrm{traj}}(\hat{\Delta}_j)\|_1},
\end{equation}
with $\hat{\Delta}_j=\Delta_j$ for MI-FGSM and the Nesterov look-ahead $\hat{\Delta}_j=\Delta_j+\alpha\mu\,g_j$ for NI-FGSM, followed by
\begin{equation}
\label{eq:update}
\Delta_{j+1}=\mathrm{Clip}_{\epsilon}\big(\Delta_j+\alpha\cdot\mathrm{sign}(g_{j+1})\big).
\end{equation}
FGSM takes a single step of size $\epsilon$. Algorithm~\ref{alg:attack} summarizes the full attack.

\textbf{Why the trajectory objective transfers.} First, Eq.~\eqref{eq:traj} depends on the environment dynamics, which the surrogate and the victim share, rather than on the surrogate's exact action at each step, so perturbations that push the system toward failure over several steps help against any policy that reacts similarly. Second, a larger temperature $\kappa$ smooths the surrogate's decision boundary, which differs between agents, and yields gradients that follow the overall trend of the Q-function, similar to the gradient stabilization of momentum~\cite{dong2018boosting}.

\begin{algorithm}[t]
\footnotesize
\caption{Trajectory-Level Transfer Attack}
\label{alg:attack}
\begin{algorithmic}[1]
\Require Surrogate Q-network $Q_s$, black-box victim policy $\pi_v$, dynamics model $f$, budget $\epsilon$, horizon $K$, iterations $T$, momentum decay $\mu$, temperature $\kappa$, operator $\mathcal{O}\in\{\text{FGSM},\text{MI},\text{NI}\}$
\Ensure Perturbation sequence that minimizes the victim's return $J$ in Eq.~\eqref{eq:goal}
\State Observe $s_0$; initialize $\Delta\gets\mathbf{0}_{K}$
\For{$t=0,1,\dots,H$}
    \State $\alpha\gets\epsilon/T$, $\Delta_0\gets\Delta$, $g_0\gets\mathbf{0}$
    \For{$j=0,\dots,T-1$}
        \State $\hat{\Delta}_j\gets\Delta_j+\alpha\mu\,g_j$ (NI) or $\Delta_j$ (MI)
        \State Roll out $f$ from $s_t$ with $\tilde{\pi}_s$ (Eq.~\eqref{eq:relax}) under $\hat{\Delta}_j$
        \State Compute $\mathcal{L}_{\mathrm{traj}}$ via Eq.~\eqref{eq:traj} and backpropagate
        \State Update $g_{j+1}$ via Eq.~\eqref{eq:mi} and $\Delta_{j+1}$ via Eq.~\eqref{eq:update}
    \EndFor \Comment{FGSM: one step of size $\epsilon$}
    \State Victim acts $a_t\gets\pi_v(s_t+\delta_t)$, $\delta_t=\Delta_T[0]$ \Comment{Transfer}
    \State Observe $r_t,s_{t+1}$; \textbf{break} if terminal
    \State $\Delta\gets$ $\Delta_T$ shifted by one step \Comment{Warm start}
\EndFor
\end{algorithmic}
\end{algorithm}

\begin{table*}[t]
\centering
\caption{Attack results on CartPole in the white-box, cross-model, and cross-algorithm settings. Each entry is the average episode return of the victim (lower means a stronger attack), reported as mean and standard deviation over the five victims (white-box), the twenty ordered surrogate--victim pairs of the same algorithm with different seeds (cross-model), or the twenty-five pairs in which the surrogate uses the other algorithm (cross-algorithm). The random baselines use uniform noise and random corner noise with the same budget and are evaluated directly on the victims. The middle attack group optimizes each step independently, and the bottom group optimizes perturbation sequences over a planning horizon. Bold marks the strongest attack in each column.}
\label{tab:results}
\setlength{\tabcolsep}{3pt}
\footnotesize
\begin{adjustbox}{max width=\textwidth}
\begin{tabular}{l|cccc|cccc|cccc}
\toprule
 & \multicolumn{4}{c}{White-box} & \multicolumn{4}{c}{Cross-model transfer} & \multicolumn{4}{c}{Cross-algorithm transfer} \\
\cmidrule(lr){2-5}\cmidrule(lr){6-9}\cmidrule(lr){10-13}
Budget & \multicolumn{2}{c}{0.05} & \multicolumn{2}{c}{0.1} & \multicolumn{2}{c}{0.05} & \multicolumn{2}{c}{0.1} & \multicolumn{2}{c}{0.05} & \multicolumn{2}{c}{0.1} \\
\cmidrule(lr){2-3}\cmidrule(lr){4-5}\cmidrule(lr){6-7}\cmidrule(lr){8-9}\cmidrule(lr){10-11}\cmidrule(lr){12-13}
Victim & DQN & DDQN & DQN & DDQN & DQN & DDQN & DQN & DDQN & DDQN & DQN & DDQN & DQN \\
\midrule
No attack & 500.0{\tiny$\pm$0} & 485.4{\tiny$\pm$24} & 500.0{\tiny$\pm$0} & 485.4{\tiny$\pm$24} & 500.0{\tiny$\pm$0} & 485.4{\tiny$\pm$24} & 500.0{\tiny$\pm$0} & 485.4{\tiny$\pm$24} & 485.4{\tiny$\pm$24} & 500.0{\tiny$\pm$0} & 485.4{\tiny$\pm$24} & 500.0{\tiny$\pm$0} \\
\midrule
Uniform noise & 375.0{\tiny$\pm$130} & 304.7{\tiny$\pm$193} & 172.7{\tiny$\pm$53} & 247.0{\tiny$\pm$193} & 375.0{\tiny$\pm$130} & 304.7{\tiny$\pm$193} & 172.7{\tiny$\pm$53} & 247.0{\tiny$\pm$193} & 304.7{\tiny$\pm$193} & 375.0{\tiny$\pm$130} & 247.0{\tiny$\pm$193} & 172.7{\tiny$\pm$53} \\
Sign noise & 240.5{\tiny$\pm$92} & 284.9{\tiny$\pm$204} & 74.3{\tiny$\pm$38} & 121.2{\tiny$\pm$133} & 240.5{\tiny$\pm$92} & 284.9{\tiny$\pm$204} & 74.3{\tiny$\pm$38} & 121.2{\tiny$\pm$133} & 284.9{\tiny$\pm$204} & 240.5{\tiny$\pm$92} & 121.2{\tiny$\pm$133} & 74.3{\tiny$\pm$38} \\
\midrule
FGSM & 322.0{\tiny$\pm$140} & 336.0{\tiny$\pm$179} & 262.5{\tiny$\pm$177} & 214.7{\tiny$\pm$163} & 330.9{\tiny$\pm$138} & 314.7{\tiny$\pm$204} & 173.4{\tiny$\pm$128} & 274.6{\tiny$\pm$185} & 302.1{\tiny$\pm$212} & 411.2{\tiny$\pm$124} & 163.4{\tiny$\pm$138} & 364.5{\tiny$\pm$140} \\
MI-FGSM & 315.8{\tiny$\pm$157} & 277.0{\tiny$\pm$200} & 159.5{\tiny$\pm$174} & 125.4{\tiny$\pm$188} & 338.9{\tiny$\pm$140} & 317.0{\tiny$\pm$209} & 147.7{\tiny$\pm$120} & 211.2{\tiny$\pm$190} & 317.7{\tiny$\pm$214} & 400.9{\tiny$\pm$127} & 148.6{\tiny$\pm$160} & 324.0{\tiny$\pm$166} \\
NI-FGSM & 304.3{\tiny$\pm$163} & 286.3{\tiny$\pm$201} & 153.5{\tiny$\pm$176} & 125.4{\tiny$\pm$189} & 360.5{\tiny$\pm$138} & 306.7{\tiny$\pm$206} & 158.2{\tiny$\pm$121} & 215.6{\tiny$\pm$189} & 319.2{\tiny$\pm$204} & 400.2{\tiny$\pm$130} & 169.7{\tiny$\pm$184} & 322.3{\tiny$\pm$164} \\
\midrule
Traj-FGSM & 113.0{\tiny$\pm$108} & 243.9{\tiny$\pm$211} & 56.5{\tiny$\pm$52} & 30.6{\tiny$\pm$23} & \textbf{190.0}{\tiny$\pm$147} & \textbf{240.0}{\tiny$\pm$205} & 77.3{\tiny$\pm$54} & 43.3{\tiny$\pm$30} & \textbf{259.1}{\tiny$\pm$214} & \textbf{181.8}{\tiny$\pm$126} & \textbf{41.6}{\tiny$\pm$27} & 82.3{\tiny$\pm$74} \\
Traj-MI-FGSM & \textbf{73.5}{\tiny$\pm$49} & \textbf{157.3}{\tiny$\pm$178} & 19.5{\tiny$\pm$13} & 15.7{\tiny$\pm$6} & 203.4{\tiny$\pm$154} & 268.1{\tiny$\pm$217} & 60.6{\tiny$\pm$95} & \textbf{41.1}{\tiny$\pm$40} & 266.9{\tiny$\pm$216} & 186.3{\tiny$\pm$162} & 47.5{\tiny$\pm$47} & \textbf{53.1}{\tiny$\pm$57} \\
Traj-NI-FGSM & 74.1{\tiny$\pm$44} & 173.1{\tiny$\pm$177} & \textbf{18.5}{\tiny$\pm$12} & \textbf{13.9}{\tiny$\pm$4} & 202.9{\tiny$\pm$149} & 271.6{\tiny$\pm$219} & \textbf{59.6}{\tiny$\pm$78} & 45.5{\tiny$\pm$40} & 268.9{\tiny$\pm$214} & 189.3{\tiny$\pm$154} & 55.1{\tiny$\pm$50} & 58.7{\tiny$\pm$62} \\
\bottomrule
\end{tabular}
\end{adjustbox}
\end{table*}

\section{Experiments}
\label{sec:exp}
\subsection{Experimental Setup}
\begin{figure}[t]
    \centering
    \includegraphics[width=0.78\columnwidth]{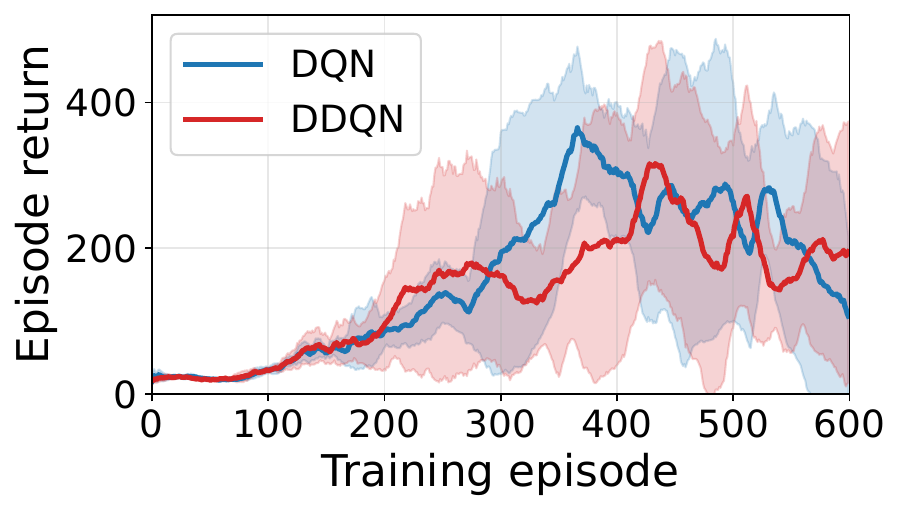}
    \caption{Training curves of the DQN and DDQN agents on CartPole. Each line is the mean episode return over five random seeds, smoothed with a 20-episode moving average, and the shaded band shows one standard deviation across seeds. The best checkpoint of each agent is selected by periodic greedy evaluation and then used as a surrogate or victim.}
    \label{fig:training}
\end{figure}

\textbf{Environment and agents.} We use CartPole-v1 from Gymnasium~\cite{brockman2016openai,6313077}. The observation is a raw 4-D vector (cart position, cart velocity, pole angle, and pole angular velocity), there are two actions (push left or right), and the agent receives a reward of $+1$ per step, up to a maximum return of 500. We train DQN~\cite{mnih2015human} and Double DQN (DDQN)~\cite{van2016deep} agents with five seeds each, which gives ten models. All agents use a two-layer MLP (128 units), Adam (learning rate $5\times10^{-4}$), batch size 128, soft target updates, and 600 training episodes, and we keep the best checkpoint under periodic greedy evaluation. Figure~\ref{fig:training} shows the training curves. Without attack, eight agents obtain the maximum return of 500, and the other two average 489.3 and 437.8.

\textbf{Attacks.} We compare the per-step FGSM, MI-FGSM, and NI-FGSM attacks (Eq.~\eqref{eq:surrogate}, lure action $a^\dagger=$ push left) with their trajectory-level counterparts Traj-FGSM, Traj-MI-FGSM, and Traj-NI-FGSM (Algorithm~\ref{alg:attack}). All iterative attacks use $T=10$ iterations and $\mu=1.0$. The trajectory attacks use horizon $K=20$ and temperature $\kappa=2$ and roll out the CartPole equations of motion. We also report two random baselines with the same budget: \emph{Uniform} noise drawn from $[-\epsilon,\epsilon]^4$, and \emph{Sign} noise that places every coordinate at a random corner, $\pm\epsilon$. Budgets are given in raw observation units, and we report $\epsilon\in\{0.05,0.1\}$.

\textbf{Protocol.} We evaluate three settings: \emph{white-box} (the surrogate is the victim), \emph{cross-model} (same algorithm, different seed; 20 ordered pairs per algorithm), and \emph{cross-algorithm} (DQN$\leftrightarrow$DDQN; 25 pairs per direction). Each evaluation runs 10 episodes with fixed environment seeds, and we report the victim's average return (lower is stronger).

\subsection{Main Results}
\textbf{Per-step attacks transfer but are weak.} Iterative per-step attacks are clearly stronger than FGSM: in the white-box setting at a budget of 0.1, MI-FGSM and NI-FGSM reduce the DQN return from 500 to 159.5 and 153.5, whereas FGSM only reaches 262.5 (Table~\ref{tab:results}, white-box). They transfer across seeds almost without loss (147.7 from one DQN to another) but are consistently weaker than Sign noise (74.3). Transfer across algorithms is also asymmetric: per-step attacks crafted on DDQN barely affect DQN victims, whose return stays above 320 (Table~\ref{tab:results}, cross-algorithm).

\textbf{Trajectory-level attacks are much stronger.} In the white-box setting at a budget of 0.1, Traj-NI-FGSM reduces the return to 18.5 on DQN and 13.9 on DDQN, 4--9$\times$ lower than Sign noise. At the smaller budget of 0.05, where all per-step attacks leave the victims above a return of 277, Traj-MI-FGSM reaches 73.5 on DQN and 157.3 on DDQN.

\textbf{Transfer across seeds and algorithms.} For cross-model transfer at a budget of 0.1, the trajectory attacks reduce DQN and DDQN victims to 59.6 and 41.1, compared with 147.7 and 211.2 for the best per-step attack and 74.3 and 121.2 for Sign noise. They also remove the asymmetry of cross-algorithm transfer: perturbations crafted on DDQN reduce DQN victims to 53.1, compared with more than 320 for per-step attacks. This supports our hypothesis that an objective defined through the shared dynamics transfers better than one defined through individual actions. At a budget of 0.05, they remain the strongest in every column, with a smaller margin (e.g., 190.0 compared with 240.5 for Sign noise).

\textbf{Momentum and look-ahead.} Momentum and look-ahead help most in the white-box setting (56.5 for Traj-FGSM compared with about 19 on DQN at a budget of 0.1). Under transfer, the three trajectory operators perform similarly, so the trajectory objective itself is the main source of transferability.

\section{Conclusion}
Per-step transfer attacks from image classification transfer across DRL agents but are no stronger than random corner noise. Optimizing perturbation sequences over a receding horizon through a differentiable environment model makes the same optimizers far more effective in white-box, cross-model, and cross-algorithm settings, which calls for defenses against temporally coordinated perturbations. Extending the attack to high-dimensional observations and learned dynamics models is left for future work.

\section*{Limitations}
Our evaluation is limited to CartPole's low-dimensional observations, known dynamics, and DQN/DDQN agents. The attack assumes access to the true state and a differentiable dynamics model; its effectiveness with visual inputs, learned or inaccurate dynamics, other environments, and adversarially trained victims remains untested. Reported standard deviations describe variability across victims or pairs, but shared models prevent treating all pairs as independent samples for significance testing. Ablations of the planning horizon, temperature, and warm start, as well as measurements of attack latency, remain future work.

\section*{Ethical Statement}
This work studies adversarial attacks to support robustness evaluation of reinforcement learning systems. The reported experiments use trained agents in simulated CartPole. The method has dual-use potential: transferable observation perturbations could be misused to disrupt deployed controllers, with possible physical harm in safety-critical applications. Such testing should therefore be restricted to systems for which explicit authorization has been obtained and conducted in simulation or controlled environments with appropriate safeguards.

\section*{Acknowledgments}
This research is supported by UDF01004676.

\bibliographystyle{plainnat}
\bibliography{custom}

\end{document}